\documentclass[letterpaper, 10pt, conference]{IEEEtran}  

\IEEEoverridecommandlockouts                              

\usepackage{times}
\usepackage{soul}
\usepackage{url}
\usepackage[hidelinks]{hyperref}
\usepackage[utf8]{inputenc}
\usepackage[small]{caption}
\usepackage{graphicx}
\usepackage{amsmath}
\usepackage{booktabs}
\usepackage{algorithm}
\usepackage{algorithmic}
\usepackage{subcaption}
\usepackage{color}
\usepackage{ifthen}
\definecolor{darkgreen}{rgb}{0,0.5,0}
\newboolean{showcomments}
\setboolean{showcomments}{true}
\newcommand{\desmond}[1]{\ifthenelse{\boolean{showcomments}}{\textcolor{red}{(Desmond says: #1)}}{}}
\newcommand{\thien}[1]{\ifthenelse{\boolean{showcomments}}{\textcolor{red}{(Thien says: #1)}}{}}
\newcommand{\shiau}[1]{\ifthenelse{\boolean{showcomments}}{\textcolor{red}{(Shiau Hong says: #1)}}{}}
\newcommand{\laura}[1]{\ifthenelse{\boolean{showcomments}}{\textcolor{red}{(Laura says: #1)}}{}}

\usepackage{amssymb}
\usepackage{mathtools}

\title{\LARGE Variational Bayesian Inference for Crowdsourcing Predictions}

\author{Desmond Cai, Duc Thien Nguyen, Shiau Hong Lim, Laura Wynter
\thanks{The authors are with IBM Research, Singapore. Emails:
\{desmond.cai1@, Duc.Thien.Nguyen@, shonglim@sg., lwynter@sg.\}ibm.com}}

\begin{document}

\maketitle

\begin{abstract}
  Crowdsourcing has emerged as an effective means for performing a number of machine learning tasks such as annotation and labelling of images and other data sets. In most early settings of crowdsourcing, the task involved classification, that is assigning one of a discrete set of labels to each task.  Recently, however, more complex tasks have been attempted including asking crowdsource workers to assign continuous labels, or predictions. In essence, this involves the use of crowdsourcing for function estimation. We are motivated by this problem to drive applications such as collaborative prediction, that is, harnessing the wisdom of the crowd to predict quantities more accurately.
  To do so, we propose a Bayesian approach aimed specifically at
   alleviating
  overfitting, a typical impediment to accurate prediction models in practice. 
  In particular, we develop a variational Bayesian technique
  for two different worker noise models -- 
  one that assumes workers' noises are independent and
  the other that assumes workers' noises have a latent low-rank structure.
  Our evaluations on synthetic and real-world datasets
  demonstrate that these Bayesian approaches  perform
  significantly better than existing non-Bayesian approaches and are thus potentially useful for this class of crowdsourcing problems.
\end{abstract}

\section{Introduction}

Crowdsourcing has proven itself as a useful and effective means for large-scale annotation of images and labelling of other machine learning data sets. Initial tasks that successfully leveraged crowdsourcing involved discrete classification tasks, such as choosing one of a small set of possible labels for each item.  Recently, more complex tasks have emerged as good candidates for crowdsourcing, including the use of continuous labels. One example is the assigning of a person's age from an image (see, for example \cite{FGNet}). More complex still is the task of asking crowdsource workers to predict a particular quantity or price, for example, the future price of a commodity. In essence, this amounts to using crowdsourcing for function estimation. We are motivated by these latter applications of collaborative prediction, that is, harnessing the wisdom of the crowd to predict quantities more accurately.

One of the fundamental questions in all forms of  crowdsourcing is how to best 
aggregate the output of the workers, be it annotations or predictions, when multiple workers perform the same task, as is often the case.
The natural method for aggregating multiple values when they are from a discrete set is majority vote. In the continuous label, or prediction, context, however, a more complex approach is needed. 
Perhaps the most common method used in this setting is inverse variance weighting. 
In that approach, the variance of each crowdsource worker 
is estimated independently. Then, a consensus label is 
computed as a weighted sum of the inverse estimated variances.  
Under the assumption of independent predictions, 
 aggregation 
based on  inverse variance makes sense. 

While inverse variance weighting offers a well-motivated form of  aggregation, the  predictions resulting from such aggregation can often overfit. In addition, the independence assumption is  violated in a number of important applications. 
For example, in many crowdsourcing contexts, 
worker predictions are based on overlapping information. 
One such example is medical text annotation 
as used for training IBM Watson components 
including Medical Relation or Factor Extraction and 
Question-Answer passage alignment \cite{watson}. 
In this case, the crowdsource workers are technical experts and 
can possess similar biases which influence their annotation errors. 
Another example is crowdsource prediction markets \cite{freeman2017crowdsourced}, 
in which case the crowdsource workers, or traders, 
can be privy to overlapping sources of information 
affecting the pattern of errors in their predictions.

We are thus interested in extending the inverse-variance approach to address these two related issues of reducing overfitting of the aggregated predictions and taking into account cases of non-independent predictions across workers.
To do so, we develop  variational Bayesian approaches for 
two different worker noise models -- 
one that assumes workers' noise are independent and 
the other that assumes workers' noises have a latent low-rank structure, thereby capturing correlations across workers.
Our evaluations on synthetic and a real-world dataset demonstrate that 
our Bayesian approaches can perform significantly better than 
existing non-Bayesian approaches and can reduce overfitting in the resulting predictions.

\section{Related work}

One of the earliest efforts in prediction aggregation was that of  \cite{dawid1979maximum} who proposed a Bayesian inference algorithm to aggregate individual worker labels and infer the ground-truth  in categorical annotation. Their approach defined the two main components needed in what was later to be known as crowdsource annotation: estimating the reliability of each crowdsource worker, and inferring the true label of each instance.  In their seminal paper, \cite{dawid1979maximum}
applied expectation maximization and estimated the ground-truth in the E-step. Then, using the
estimated ground-truth, they compute the maximum likelihood estimates of the confusion matrix
in the M-step.

In continuous value annotation, more recently, \cite{raykar2010learning} modeled each worker prediction as an independent noisy observation of the ground-truth. Based on this independent noise assumption, \cite{raykar2010learning} developed a counterpart  to the  Dawid-Skene framework for the continuous domain to infer both the unknown individual variance and the ground-truth.   In their M-step, the variance, which  corresponds to the confusion matrix in categorical annotation, is computed to minimize the mean square error with respect to the estimated ground-truth. Their  E-step involves re-estimating the ground-truth with a weighted sum of the individual predictions, where  the weights are set as  the inverses of individual variances. 
The authors of \cite{liu_variational} point to the risk of convergence to a poor-quality local optimum of the above-mentioned EM approaches and instead propose a variational approach for the problem. 
 Later, \cite{liu2013scoring} took a different tack by injecting  items with ground-truth labels, called control questions, to better estimate the reliability of each crowdsource worker in what they term their ``consensus model with partial ground truth''. Also related to our work is the work of \cite{zhou2014aggregating} to extend confusion matrix in categorical classification into ordinal label aggregation. However, as the focus of our work is to develop continuous value annotation aggregation, we leave the comparison with ordinal label methods for future work.

The authors of \cite{NIPS2010_4074}  seek to define groups of behaviors by modeling each crowdsource worker as a multi-dimensional quantity including bias and other factors, and then group them as a function of those quantities.
 In a similar vein, \cite{kara2015modeling} aim to identify the particular types of worker behavior that they wish to model, including bias and 
 the so-called worker opinion scaling. A common technique to aggregate annotations in crowdsourcing regression is to use a weighted sum with the weights proportional to the confidence in each workers' prediction, as in  \cite{liu2013scoring,raykar2010learning,li2014confidence}. However, such a weighted sum  does not account for potential  correlation between workers. 
Also in a Bayesian setting and also motivated by the Dawid-Skene framework, \cite{ok2017iterative} propose an iterative Bayesian update but rely on a discretization of the crowdsource workers and their variances into a small number of classes.

Crowdsourcing data  are typically sparse  as  the workers seldom participate in all of the tasks. This  sparsity  motivates  the application of collaborative filtering and matrix factorization techniques to the crowdsourcing domain. Along these lines, \cite{zhang2014spectral} proposed a spectral method to  provide an initial estimate of the confusion matrix for an EM algorithm. Another application of matrix factorization was proposed by \cite{jung2012improving} to infer the unobserved predictions of workers before applying standard aggregation methods, i.e. majority voting. In \cite{zhou2016crowdsourcing}, a method was devised to augment the ground-truth   by using low-rank tensor completion to infer  the missing entries of the label matrix. An advantage of the above-mentioned matrix factorization  approaches is that they model the correlation of  errors for  each individual crowdsource worker across instances, as represented by the confusion matrix. 

 However, in many cases, especially the continuous labeling applications such as the medical domain and prediction markets that interest us, crowdsource workers may exhibit correlations in their responses across the workers themselves.   In \cite{li2019exploiting} it was shown that the incorporation of cross-worker correlations significantly improves accuracy. Their work, which  focused on  the discrete problem of unsupervised crowdsourced classification, relies on an extension of the (independent) Bayesian Classifier Combination model of \cite{iBCC} in which they model worker correlation by considering that  true classes
are mixtures of subtypes.

Applications in the continuous-label context are often equipped with  some ground truth labels on a subset of the data. This occurs not only when control questions are injected for the purpose of enhancing the model accuracy for each worker, as described by \cite{liu2013scoring}, but also in applications where ground truth is obtained naturally later in time, such as in medical diagnosis and prediction markets.

It is in this context that  \cite{volkovs2012learning}  propose a model to aggregate the rankings of crowdsource workers, or ``experts''.
They represent the ranking matrix of each  worker by  a low-rank decomposition so as to learn the ranking regression function for the  aggregation.  The supervised framework that they propose is not directly applicable to our setting as  it assumes a fixed set of $K$ crowdsource workers across all instances. In addition, their method of concatenating   experts'  features leads to a model of rather high dimension,  not suitable for the regression tasks that interest us here.

\section{Maximum Likelihood Estimation}

We can state  the problem formally as follows: we seek to estimate the ground truth values 
for a set of $I$ regression outputs  denoted by $i = 1,\dots,I$.
The regression output is composed of predictions provided 
by a set of $J$ crowdsource workers, themselved denoted by $j = 1,\dots,J$.

Each crowdsource worker  can participate in a subset of the prediction tasks.
Let $J_i \subseteq \{1,\dots,J\}$ denote the set of workers that participated in task $i$
and let $I_j \subseteq \{1,\dots,I\}$ denote the set of tasks that were assigned to worker $j$.
Let $y_i\in R$ denote the ground truth value for task $i \in I$
and $x_{ij}\in R$ denote the prediction provided by worker $j \in J_i$.
In summary, the goal is to estimate the ground truth  from the available predictions.

\subsection{Inverse Variance Weighting}

One of the most popular approaches for aggregating predictions 
is the inverse variance weighting method. 
Inverse variance weighting has an appealing interpretation as maximum-likelihood estimation 
under the bias-variance model. 
The latter  is based on the assumption that 
workers have independent additive prediction noise
~\cite{liu2013scoring,raykar2010learning,kara2015modeling}:
\begin{equation*}
  p\left(x_{ij}\left|y_i,\sigma_j^2\right.\right)
  = 
  \mathcal{N}\left(x_{ij}|y_i,\sigma_j^2\right).
  \label{eq:bias-variance-model}
\end{equation*}
That is, the noise is assumed to be Gaussian with variance $\sigma_j^2$.
The inverse variance weighting method finds 
the ground truth values $\mathbf{y}=(y_1,\dots,y_I)$ 
and noise variance $\boldsymbol\sigma^2=(\sigma_1^2,\dots,\sigma_J^2)$
that maximize the log-likelihood of the predictions:
\begin{align*}
  \log p\left(\mathbf{X}\left|\mathbf{y},\boldsymbol\sigma^2\right.\right)
  =
  \sum_{(i,j): i\in I_j} \log p\left(x_{ij}\left|y_{i}, \sigma_j^2\right.\right).
\end{align*}

One approach to maximize the log-likelihood is to perform block coordinate descent 
by alternating between the optimisation over $\boldsymbol{y}$ and $\boldsymbol{\sigma}^2$:
\begin{subequations}
\begin{align}
  y_i
  &= 
  \frac{\sum_{j\in J_i}\frac{1}{\sigma_j^2}x_{ij}}{\sum_{j\in J_i}\frac{1}{\sigma_j^2}},
  \label{eq:ivar:1}
  \\
  \sigma_j^2
  &= 
  \frac{1}{|I_j|}\sum_{i\in I_j}\left(x_{ij} - y_i\right)^2.
  \label{eq:ivar:2}
\end{align}
\label{eq:ivar}%
\end{subequations}
This simple procedure has been observed to perform well  when 
worker predictions are independently generated and there is little overlap in 
their information and methods.

\subsection{Inverse Covariance Weighting}
\label{sec:icov}

The independence assumption in the bias-variance model 
is  violated in a number of important, real-world crowdsourcing scenarios. 
Specifically, when workers  use similar information and methods to make their predictions,
this leads to a collective bias within groups or sub-groups of participants.
We therefore require a  method that can exploit these
correlations between the prediction noise of different workers.

Let $\mathbf{j}_i\in R^{|J_i|}$ denote the vector of indices of workers 
that provided predictions for task $i$. 
Denote by $x_{i,\mathbf{j}_i}\in R^{|J_i|}$ the 
subvector of observed predictions for task $i$.
One straightforward extension is to consider a 
multivariate Gaussian model with covariance $\boldsymbol\Sigma$ 
for the prediction noise:
\begin{equation}
  p\left(x_{i,\mathbf{j}_i}\left|y_i,\boldsymbol\Sigma_{\mathbf{j}_i}\right.\right)
  = 
  \mathcal{N}\left(x_{i,\mathbf{j}_i}\left|y_i\mathbf{1},\boldsymbol\Sigma_{\mathbf{j}_i}\right.\right),
  \label{eq:bias-covariance-model}
\end{equation}
where $\mathbf{1}$ is a vector of $1$'s of the appropriate dimension
and $\boldsymbol\Sigma_{\mathbf{j}_i} \in R^{|J_i|\times|J_i|}$ 
is the submatrix of $\boldsymbol\Sigma$ containing the rows and columns
corresponding to the indices in $\mathbf{j}_i$.
Similar to the inverse variance weighting method, 
block coordinate descent can be used to maximize the log-likelihood 
by alternating between the optimisation over $\mathbf{y}$ and $\boldsymbol\Sigma$
as follows:
\begin{subequations}
  \begin{align}
    y_i
    &= 
    \frac{\mathbf{1}^\top {\boldsymbol\Sigma}_{\mathbf{j}_i}^{-1} \mathbf{x}_{\mathbf{j}_i}}
    {\mathbf{1}^\top {\boldsymbol\Sigma}_{\mathbf{j}_i}^{-1}\mathbf{1}},
    \label{eq:icov:1}
    \\
    \Sigma_{jj'}
    &=
    \frac{1}{|I_j \cap I_{j'}|}
    \sum_{i \in I_j\cap I_{j'}} 
    \left(\delta_{ij} - \bar{\delta}_j\right)
    \left(\delta_{ij'} - \bar{\delta}_{j'}\right),
    \label{eq:icov:2}
  \end{align}
  \label{eq:icov}%
\end{subequations}
where $\delta_{ij}\coloneqq x_{ij} - y_i$ is 
the prediction error of worker $j$ on task $i$
and $\bar{\delta}_j\coloneqq (1/|I_j|)\sum_{i\in I_j}\delta_{ij}$ is 
the mean prediction error of worker $j$.
Note that $\boldsymbol\Sigma$ contains $|J|^2$ elements and, 
in equation~\eqref{eq:icov:2}, each element $\Sigma_{jj'}$ 
is simply updated with the sample covariance between workers $j$ and $j'$.
In order for this procedure to be accurate,
there must be a substantial number of common tasks between both workers, $i$ and $j$. 
However, in crowdsourcing applications, it is typically the case that observations are sparse, and as such this approach is  unlikely to work well in practice.

This motivates the consideration of a latent feature noise model that
allows for correlations between the prediction noise of different workers
while also addressing the challenge of  sparse observations.
In particular, we consider the following probabilistic model for the workers:
\begin{equation*}
  p\left(x_{ij}\left|y_i,u_i,v_j,\sigma^2\right.\right)
  =
  \mathcal{N}\left(x_{ij}\left|y_i+u_i^\top v_j,\sigma^2\right.\right),
\end{equation*}
where $u_i\in R^D$ and $v_j\in R^D$ are latent noise feature vectors 
associated with task $i$ and worker $j$ respectively 
and the noise is assumed to be Gaussian with variance $\sigma^2$.

The problem of inferring the latent noise feature vectors 
by maximizing the log-likelihood 
resembles classical matrix factorization problems 
(the only difference is the presence of an additive ground truth term $y_i$).
In the rest of this section, we tackle the problem of 
maximizing the log-likelihood:
\begin{align*}
  \log p\left(\mathbf{X}\left|\mathbf{y},\mathbf{U},\mathbf{V},\boldsymbol\sigma^2\right.\right)
  =
\sum_{(i,j): i\in I_j} \log p\left(x_{ij}\left|y_{i}, \mathbf{u}_i, \mathbf{v}_j,\sigma^2\right.\right),
\end{align*}
where the matrices 
$\mathbf{U}\coloneqq \left[\mathbf{u}_1,\dots,\mathbf{u}_I\right]^\top \in R^{I\times D}$ and 
$\mathbf{V}\coloneqq \left[\mathbf{v}_1,\dots,\mathbf{v}_J\right]^\top \in R^{J\times D}$.
In particular, we extend the inverse covariance weighting method 
with a nonlinear matrix factorization technique 
based on Gaussian processes~\cite{lawrence2009non}
to jointly infer the ground truth values and latent noise feature vectors.

Observe that, by placing independent zero mean Gaussian priors 
$\mathcal{N}(\mathbf{0},\sigma_u^2\mathbf{I})$ on $\mathbf{u}_i$,
we recover our initial probabilistic model in~\eqref{eq:bias-covariance-model}
with the covariance matrix:
\begin{equation*}
\boldsymbol\Sigma = \sigma_u^2 \mathbf{V}\mathbf{V}^\top + \sigma^2\mathbf{I}.
\end{equation*}
Therefore, the problem of covariance estimation 
has been transformed into an estimation of $\mathbf{V}$, $\sigma_u^2$, $\sigma^2$.
The degrees of freedom are now primarily 
determined by the size of $\mathbf{V}$ which contains $I\times D$ values.
Since we expect $D \ll I$ in practical applications, this problem 
has significantly fewer degrees of freedom than the original problem 
of estimating the $I^2$ values of the entire covariance matrix.

To maximize the log-likelihood, we alternate between the
optimisation of $\mathbf{y}$ and $(\mathbf{V}, \sigma^2, \sigma_u^2)$. 
Specifically, we update $\mathbf{y}$ using equation~\eqref{eq:icov:1} 
and perform stochastic gradient descent on the model parameters as there 
is no closed-form solution for the latter.
The log-likelihood of task $t$ is:
\begin{align*}
  E_t(\mathbf{V},\sigma^2,\sigma_u^2)
  =
  - \log\left|\boldsymbol{\Sigma}_{\mathbf{j}_t}\right| 
  - \boldsymbol{\delta}_{\mathbf{j}_i,i}^\top \boldsymbol{\Sigma}_{\mathbf{j}_i}^{-1} \boldsymbol{\delta}_{\mathbf{j}_i,i} + \text{const.}
\end{align*}
and the gradients with respect to the parameters are:
\begin{subequations}
\begin{align}
  \nabla_{\mathbf{V}_{\mathbf{j}_i,:}} E_t(\mathbf{V},\sigma^2,\sigma_u^2)
  &=
  \mathbf{G}_i \mathbf{V}_{\mathbf{j}_i,:},
  \label{eq:sgd:feature}
  \\
  \nabla_{\sigma^2} E_t(\mathbf{V},\sigma^2,\sigma_u^2)
  &=
  \mathrm{Tr}
  \left( 
    \mathbf{G}_i
  \right),
  \label{eq:sgd:param1}
  \\
  \nabla_{\sigma_u^2} E_t(\mathbf{V},\sigma^2,\sigma_u^2)
  &=
  -
  \sigma_u^4
  \mathrm{Tr}
  \left(
    \mathbf{G}_t
    \mathbf{V}_{\mathbf{j}_i,:}
    \mathbf{V}_{\mathbf{j}_i,:}^\top
  \right).
  \label{eq:sgd:param2}
\end{align}
\label{eq:sgd}%
\end{subequations}
where 
$\mathbf{G}_i \coloneqq 
\boldsymbol{\Sigma}_{\mathbf{j}_i}^{-1} 
\boldsymbol{\delta}_{\mathbf{j}_i,i} 
\boldsymbol{\delta}_{\mathbf{j}_i,i}^\top 
\boldsymbol{\Sigma}_{\mathbf{j}_i}^{-1} - 
\boldsymbol{\Sigma}_{\mathbf{j}_i}^{-1}$ and
$\mathbf{V}_{\mathbf{j}_i,:} \in R^{|J_i| \times D}$ 
is the submatrix of $\mathbf{V}$ containing 
the rows corresponding to the indices in $\mathbf{j}_i$.
When partial ground truth information is available,
$\mathbf{y}$ would be updated only for instances without such ground truth values.
After inferring the covariance matrix, 
predicting the ground truth values for new instances can be done
by applying equation~\eqref{eq:icov:1}.

We can also model the covariance matrix with non-linear kernel functions 
by replacing the inner products 
$\mathbf{v}_j^\top \mathbf{v}_{j'}$ in the covariance expression 
by a Mercer kernel function $k(\mathbf{v}_j, \mathbf{v}_{j'})$. 
The parameters in the kernel representation can be optimized by 
gradient descent on the log-likelihood function.
In this paper, we focus on the linear kernel 
$k(\mathbf{v}_j, \mathbf{v}_{j'}) = \mathbf{v}_j^\top \mathbf{v}_{j'} $
and leave the exploration of alternative non-linear kernels to future work.

\subsection{Multidimensional Targets}

The use of inverse covariance weighting approach also provides a way to leverage correlations 
 when tasks involve multiple ground truth targets.
For example, suppose each task $i$ has a multidimensional output comprising $K$ ground truth values  
$\tilde{\mathbf{y}}_i = (y_{i1},\dots,y_{iK})$ and
worker $j$ makes a prediction 
$\tilde{\mathbf{x}}_{ij} = (x_{ij1},\dots,x_{ijK})$ on that  task $i$.
Then for each such task $i$, stack the prediction vectors to form
$\tilde{\mathbf{x}}_{i} = (\tilde{\mathbf{x}}_{i1},\dots,\tilde{\mathbf{x}}_{iK})$
and let $\tilde{\mathbf{x}}_{i,\mathbf{j}_i}$ denote the subvector 
containing only the predictions of the workers assigned to task $i$.
Applying the bias covariance model in equation~\eqref{eq:bias-covariance-model} 
to the stacked vector $\tilde{\mathbf{x}}_{i,\mathbf{j}_i}$ 
provides an approach that learns and exploits 
latent relationships between targets.

\section{Variational Bayesian Inference}

 We now present our proposed Bayesian approaches for aggregating predictions.
That is, rather than maximize the likelihood of the observations, 
we wish to compute the posterior $p(\mathbf{y}|\mathbf{X})$. 
Since this problem is typically analytically intractable,
we use variational Bayesian approximation techniques 
to develop separate approaches for approximating the posterior 
under both the independent noise and latent noise models. 

\subsection{Algorithm for Independent Noise Model}

We place independent zero mean Gaussian priors 
on the ground truth labels in the bias-variance model:
\begin{align*}
  p\left(y_i,\tau_{i}^2\right) 
  &= \mathcal{N}\left(y_i\left|0,\tau_{i}^2\right.\right),
\end{align*}
where $\tau_{i}^2$ are hyperparameters.
Note that we assumed a zero mean prior to simplify the derivations, 
implicitly assuming that the ground truth labels 
have been shifted to have zero mean,
but our procedure can be easily extended to use nonzero mean priors.

The variational approximate inference procedure 
approximates the posterior $p(\mathbf{y}|\mathbf{X})$.
by finding the distribution $q_y$
that maximize the variational free energy of the model:
\begin{equation*}
  F\left(q_y\right)
  =
  \mathbb{E}_{q_y}
  \left[
    \frac{\log p\left(\mathbf{X},\mathbf{y}\right)}
    {\log\left(q_y(\mathbf{y})\right)}
  \right],
\end{equation*}
where the joint probability:
\begin{equation*}
  p\left(\mathbf{X},\mathbf{y}\right)
  =
  \prod_{(i,j):i\in I_j}
  p\left(x_{ij}\left| y_i\right.\right)\prod_i p\left(y_i\right).
\end{equation*}
Taking the derivative of $F$ with respect to $q_y$ and setting it to zero 
implies that the stationary distributions are independent Gaussians:
\begin{equation*}
  q_y\left(\mathbf{y}\right)
  =
  \prod_i
  \mathcal{N}\left(y_i\left|\bar{y}_i,\lambda_i\right.\right).
\end{equation*}
where the means and covariances satisfy the following conditions:
\begin{align}
  \lambda_i
  &=
  \left(
    \frac{1}{\tau_i^2}
    + \sum_{j\in J_i}\frac{1}{\sigma_j^2}
  \right)^{-1},
  \label{eq:ind-vb-truth-var}
  \\
  \bar{y}_{i}
  &=
  \lambda_i 
  \sum_{j\in J_i} \frac{1}{\sigma_j^2} y_{ij}.
  \label{eq:ind-vb-truth-mean}
\end{align}
Equations~\eqref{eq:ind-vb-truth-var} and~\eqref{eq:ind-vb-truth-mean} 
provide update equations for performing 
block coordinate descent on the means and covariances.
We also update the hyperparameters using block coordinate descent. 
Differentiating $F$ and setting the derivatives to zero 
give the following updates:
\begin{align}
  \tau_j^2
  &=
  \frac{1}{|I_j|}
  \sum_{i\in I_j} \left( \lambda_i + \bar{y}_{i}^2\right),
  \label{eq:ind-vb-truth-param}
  \\
  \sigma_j^2
  &=
  \frac{1}{|I_j|}
  \sum_{i\in I_j} 
  \left(
    \lambda_i + \left(x_{ij} - \bar{y}_{i}\right)^2 
  \right).
  \label{eq:ind-vb-error-param}
\end{align}
In summary, we apply equations~\eqref{eq:ind-vb-truth-var} to~\eqref{eq:ind-vb-error-param} 
repeatedly, and perform aggregation using the posterior mean $\bar{y}_i$.

It is, in fact, tractable to compute the exact posterior 
of the ground truth labels under the bias-variance model 
after placing independent Gaussian priors on the ground truth labels.
Specifically, the posterior turns out to be independent Gaussian distributions,
that are equivalent to that computed
by equations~\eqref{eq:ind-vb-truth-var} and~\eqref{eq:ind-vb-truth-mean}.

However, computing the posterior alone 
does not provide a way to update the variance hyperparameters.
In our approach, the hyperparameters are updated by 
maximizing the variational free energy of the model which can be
interpreted as the expectation of the log joint distribution
over the posterior and is a lower bound on the log-likelihood.

\subsection{Algorithm for the  Latent Noise Model}

One of the key steps in the procedure developed in Section~\ref{sec:icov} 
is the marginalization of $\mathbf{U}$ 
conditioned on $(\mathbf{V},\sigma^2,\sigma_u^2)$. 
The marginalization step can in fact be interpreted as 
performing Bayesian averaging over $\mathbf{U}$. 
However, it turns out to be challenging to perform 
full Bayesian averaging over both $\mathbf{U}$ and $\mathbf{V}$,
which motivates us to develop a variational Bayesian approach.

First, we place independent zero mean Gaussian priors 
on the inference variables as follows:
\begin{align*}
  p\left(y_i,\sigma_y^2\right) 
  &= \mathcal{N}\left(y_i\left|0,\sigma_y^2\right.\right),
  \\
  p\left(\mathbf{u}_i,\sigma_u^2\right) 
  &= \mathcal{N}\left(\mathbf{u}_i\left|\mathbf{0},\sigma_u^2\mathbf{I}\right.\right),
  \\
  p\left(\mathbf{v}_j,\sigma_v^2\right) 
  &= \mathcal{N}\left(\mathbf{v}_j\left|\mathbf{0},\sigma_v^2\mathbf{I}\right.\right),
\end{align*}
where $\sigma_y^2$, $\sigma_u^2$, $\sigma_v^2$ are hyperparameters.
For notational brevity, we will omit the dependence of the distributions 
on the hyperparameters $\sigma_y^2$, $\sigma_u^2$, $\sigma_v^2$ 
in the rest of this section.

By conditioning on $\mathbf{U}$ and $\mathbf{V}$ and 
using variational approximate inference 
to approximate the conditional posterior, we obtain:
\begin{align*}
  p\left(\mathbf{y}\left|\mathbf{X}\right.\right)
  &=
  \mathbb{E}_{\mathbf{U},\mathbf{V}|\mathbf{X}}
  \left[
  p\left(\mathbf{y},\mathbf{U},\mathbf{V}\left|\mathbf{X}\right.\right)
  \right]
  \\
  &\approx
  \mathbb{E}_{\mathbf{U},\mathbf{V}|\mathbf{X}}
  \left[
    q_y\left(\mathbf{y}\right)
    q_u\left(\mathbf{U}\right)
    q_v\left(\mathbf{V}\right)
  \right]
  \\
  &=
  q_y\left(\mathbf{y}\right),
\end{align*}
and therefore, inference reduces to computing the mean 
with respect to the distribution $q_y$.
The variational approximate inference procedure finds distributions 
that maximize the variational free energy of the model:
\begin{equation*}
  F\left(q_y,q_u,q_v\right)
  =
  \mathbb{E}_{q_y,q_u,q_v}
  \left[
    \frac{\log p\left(\mathbf{X},\mathbf{y},\mathbf{U},\mathbf{V}\right)}
    {\log\left(q_y(\mathbf{y}),q_u(\mathbf{U})q_v(\mathbf{V})\right)}
  \right],
\end{equation*}
where the joint probability:
\begin{align*}
  p\left(\mathbf{X},\mathbf{y},\mathbf{U},\mathbf{V}\right)
  &=
  \prod_{(i,j):i\in I_j}
  p\left(x_{ij}\left| y_i,\mathbf{u}_i,\mathbf{v}_j\right.\right)
  \\
  &\qquad\;
  \times \prod_i p\left(y_i\right)
  \prod_i p\left(\mathbf{u}_i\right)
  \prod_j p\left(\mathbf{v}_j\right).
\end{align*}

We solve for $q_y$, $q_u$, $q_v$ by performing 
block coordinate descent on $F$. 
The first-order conditions imply that the stationary distributions
are independent Gaussians:
\begin{align*}
  q_y\left(\mathbf{y}\right)
  &=
  \prod_i
  \mathcal{N}\left(y_i\left|\bar{y}_i,\lambda_i\right.\right),
  \\
  q_u\left(\mathbf{U}\right)
  &=
  \prod_i
  \mathcal{N}\left(\mathbf{u}_i\left|\bar{\mathbf{u}}_i,\boldsymbol\Phi_i\right.\right),
  \\
  q_v\left(\mathbf{V}\right)
  &=
  \prod_j
  \mathcal{N}\left(\mathbf{v}_j\left|\bar{\mathbf{v}}_j,\boldsymbol\Psi_j\right.\right),
\end{align*}
where the means and covariances are given by:
\begin{align}
  \lambda_i
  &=
  \left(
    \frac{1}{\sigma_y^2}
    + \sum_{j\in J_i}\frac{1}{\sigma^2}
  \right)^{-1},
  \label{eq:latent-vb-truth-var}
  \\
  \bar{y}_{i}
  &=
  \lambda_i 
  \sum_{j\in J_i} \frac{1}{\sigma^2} \left(y_{ij} - \bar{\mathbf{u}}_i^\top \bar{\mathbf{v}}_j\right),
  \label{eq:latent-vb-truth-mean}
  \\
  \boldsymbol\Phi_i
  &=
  \left(
    \frac{1}{\sigma_u^2}\mathbf{I}
    + \frac{1}{\sigma^2}\sum_{j\in J_i}
    \left(\boldsymbol\Psi_j + \bar{\mathbf{v}}_j\bar{\mathbf{v}}_j^\top\right)
  \right)^{-1},
  \label{eq:latent-vb-worker-var}
  \\
  \bar{\mathbf{u}}_i 
  &= 
  \boldsymbol\Phi_i 
  \sum_{j\in J_i} \frac{1}{\sigma^2}\left(x_{ij} - \bar{y}_{i}\right) \bar{\mathbf{v}}_j,
  \label{eq:latent-vb-worker-mean}
  \\
  \boldsymbol\Psi_j
  &=\left(
    \frac{1}{\sigma_v^2} \mathbf{I} 
  + \sum_{i \in I_j} \frac{1}{\sigma^2}
  \left(\boldsymbol\Phi_i + \bar{\mathbf{u}}_i\bar{\mathbf{u}}_i^\top\right)
  \right)^{-1},
  \label{eq:latent-vb-task-var}
  \\
  \bar{\mathbf{v}}_j
  &= 
  \boldsymbol{\Psi}_j
  \sum_{i\in I_j}\frac{1}{\sigma^2}\left(x_{ij} - \bar{y}_{i}\right)\bar{\mathbf{u}}_i.
  \label{eq:latent-vb-task-mean}
\end{align}
The hyperparameter updates can be obtained similarly by 
differentiating $F$ and setting the derivatives to zero: 
\begin{align}
  \sigma_y^2
  &=
  \frac{1}{I}
  \left(\sum_i \left( \lambda_i + \bar{y}_{i}^2\right) \right),
  \label{eq:latent-vb-truth-param}
  \\
  \sigma_u^2
  &=
  \frac{1}{DI}
  \left(\sum_i \mathrm{Tr}\left(\boldsymbol\Phi_i + \bar{\mathbf{u}}_i \bar{\mathbf{u}}_i^\top \right)\right),
  \label{eq:latent-vb-worker-param}
  \\
  \sigma_v^2
  &=
  \frac{1}{DJ}
  \left(\sum_j \mathrm{Tr}\left(\boldsymbol\Psi_j + \bar{\mathbf{v}}_j \bar{\mathbf{v}}_j^\top \right)\right),
  \label{eq:latent-vb-task-param}
  \\
  \sigma^2
  &=
  \frac{1}{\sum_{j} |I_j|}
  \sum_{(i,j):i\in I_j} \left[
    \lambda_i + \left(x_{ij} - \bar{y}_{i}\right)^2 
    - 2 \left(x_{ij} - \bar{y}_{i}\right) \bar{\mathbf{u}}_i^\top \bar{\mathbf{v}}_j \right.
  \nonumber
  \\
  & \qquad\qquad\quad\; 
    + \mathrm{Tr}\left(
    \left(\boldsymbol\Psi_i + \bar{\mathbf{u}}_i\bar{\mathbf{u}}_i^\top\right)
    \left(\boldsymbol\Phi_j + \bar{\mathbf{v}}_j\bar{\mathbf{v}}_j^\top\right)
  \right)
\Big].
  \label{eq:latent-vb-error-param}
\end{align}
In summary, the algorithm applies 
equations~\eqref{eq:latent-vb-truth-var} to~\eqref{eq:latent-vb-error-param} 
repeatedly until convergence. 
If there are multiple targets, the above derivations can be repeated to yield 
update equations for the posterior distributions of different targets and 
their associated latent feature vectors. 

Note that, unlike in inverse covariance weighting in which prediction can be 
performed in a single step after estimating the covariance matrix, the 
variational algorithm always requires multiple iterations for prediction 
since it has to estimate the posterior latent distributions associated with each 
new instance.
In applications where this is not feasible, 
one approximate solution is to fix the hyperparameters 
and worker latent distributions $q_{v}$ 
with those obtained during training and only update 
$q_{y}$ and $q_{u}$. 
In this case, 
equations~\eqref{eq:latent-vb-truth-var},~\eqref{eq:latent-vb-truth-mean},~\eqref{eq:latent-vb-task-var},~\eqref{eq:latent-vb-task-mean}
can be solved to obtain a closed-form solution for prediction:
\begin{equation*}
  \bar{y}_{i}
  =
  \frac{\sum_{j\in J_i} \rho_{ij} x_{ij}}{
    \frac{1}{\sigma_y^2} + \sum_{j\in J_i} \rho_{ij}
  },
\end{equation*}
where:
\begin{equation*}
  \rho_{ij}
  \coloneqq
  \frac{1}{\sigma^2}
  \left(
    1 - 
    \left(\sum_{j'\in J_i} \frac{1}{\sigma^2} \bar{\mathbf{v}}_{j'}\right)^\top
    \boldsymbol\Phi_i \bar{\mathbf{v}}_j
  \right).
\end{equation*}
Another approximate solution is to use the obtained
posterior mean latent feature vectors $\mathbf{v}_j$ 
to form the covariance matrix and apply the inverse covariance method.

\section{Experimental Results}

\newcommand{\average}{$\mathsf{Average}$}
\newcommand{\indml}{$\mathsf{Ind}$-$\mathsf{ML}$}
\newcommand{\indvb}{$\mathsf{Ind}$-$\mathsf{VB}$}
\newcommand{\latentml}{$\mathsf{Latent}$-$\mathsf{ML}$}
\newcommand{\latentvb}{$\mathsf{Latent}$-$\mathsf{VB}$}

We compare the following five algorithms:
\begin{itemize}
  \item \average\, takes the average of the observed predictions
    as the aggregated value.
  \item \indml\, is the classical inverse variance weighting algorithm that 
    recursively updates ground truth estimates and worker variances 
    using equation~\eqref{eq:ivar}.
  \item \indvb\, is the Variational Bayes approach
    based on the independent noise model that 
    recursively updates ground truth posteriors and hyperparameters
    using equations~\eqref{eq:ind-vb-truth-var}~--~\eqref{eq:ind-vb-error-param}.
  \item \latentml\, is the inverse covariance weighting algorithm with a
    latent noise feature model that recursively 
    updates ground truth estimates and worker variances using equation~\eqref{eq:icov}
    and performs stochastic gradient descent on the latent features and hyperparameters
    using equation~\eqref{eq:sgd}.
  \item \latentvb\, is the Variational Bayes approach
    based on the latent noise model that recursively
    updates ground truth and latent posteriors and hyperparameters
    using equations~\eqref{eq:latent-vb-truth-var}~--~\eqref{eq:latent-vb-error-param}.
\end{itemize}

\subsection{Synthetic Datasets}

\begin{figure*}[!tb]
	\centering
	\begin{subfigure}{0.24\textwidth}
		\includegraphics[width=\textwidth]{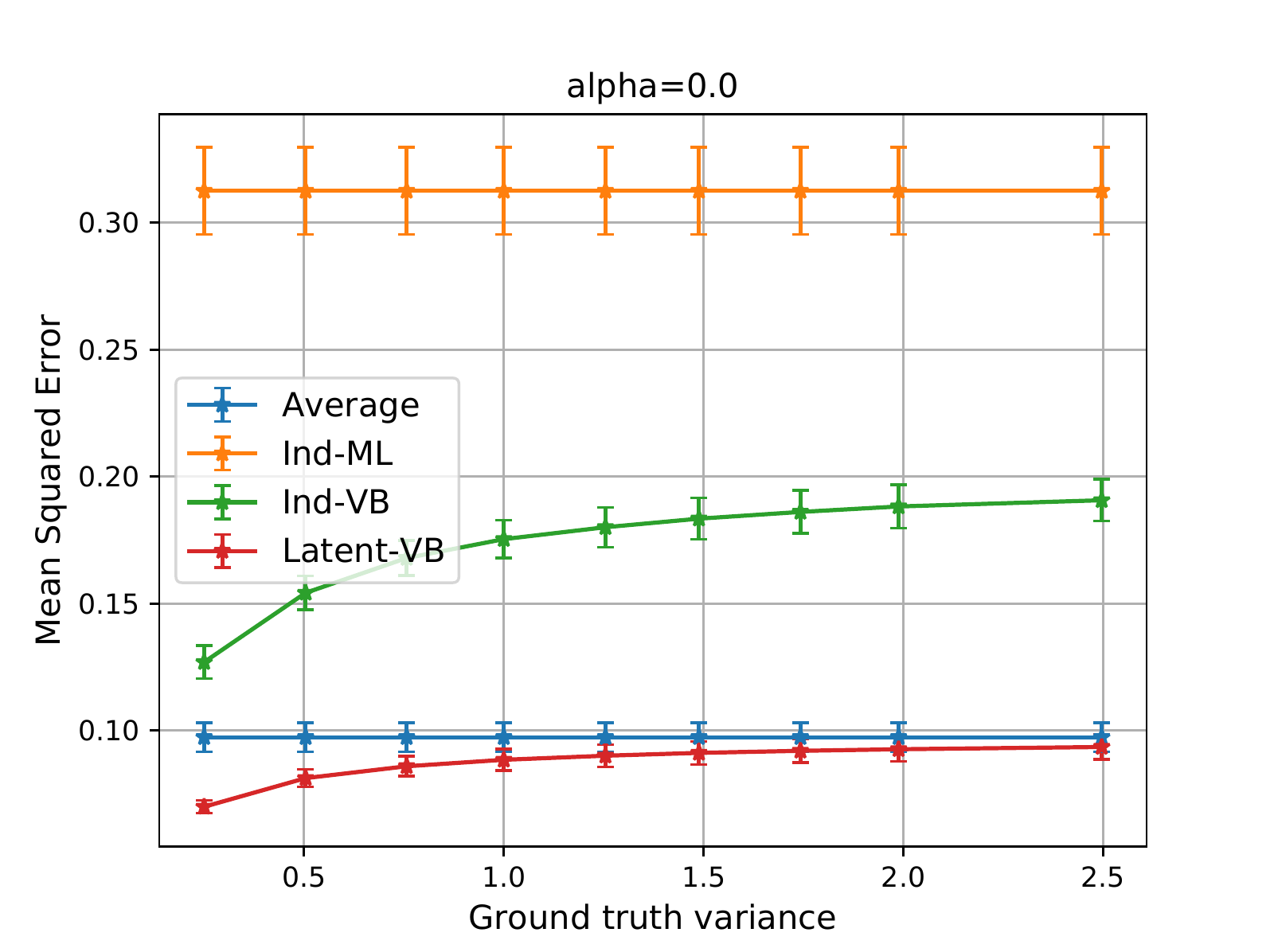} \caption{}
	\end{subfigure} 
	\begin{subfigure}{0.24\textwidth}
		\includegraphics[width=\textwidth]{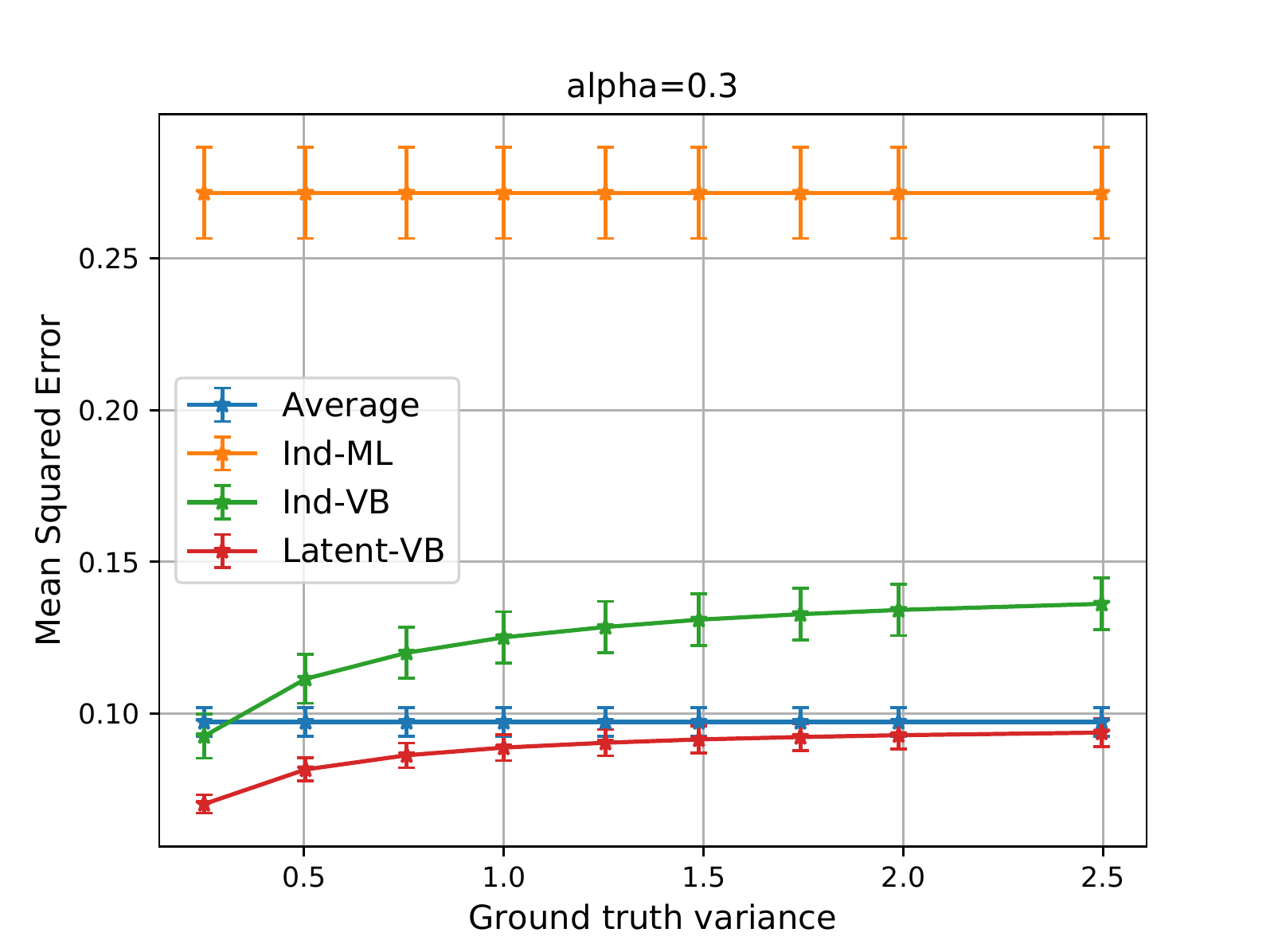} \caption{}
	\end{subfigure} 
	\begin{subfigure}{0.24\textwidth}
		\includegraphics[width=\textwidth]{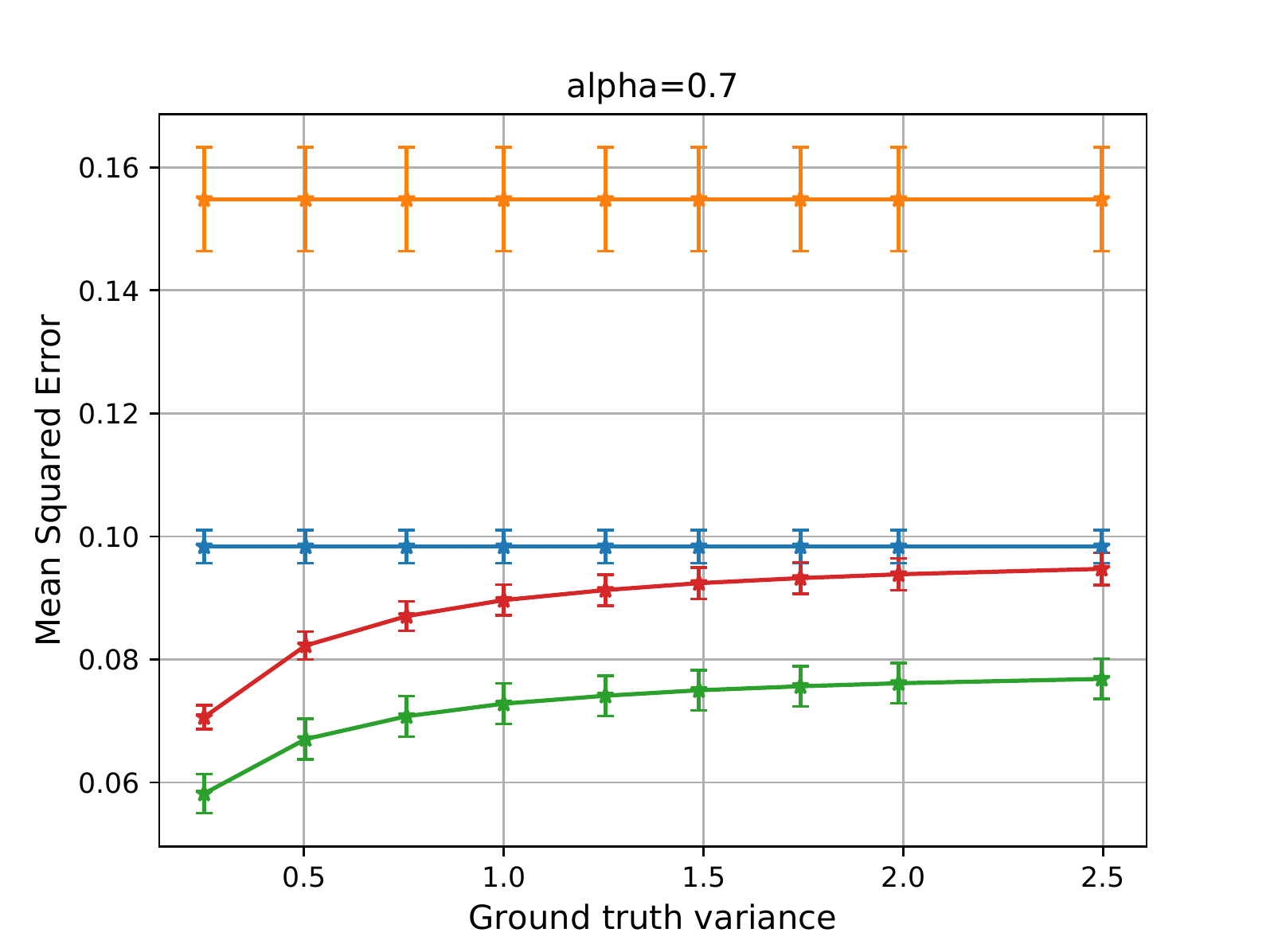} \caption{}
	\end{subfigure} 
	\begin{subfigure}{0.24\textwidth}
		\includegraphics[width=\textwidth]{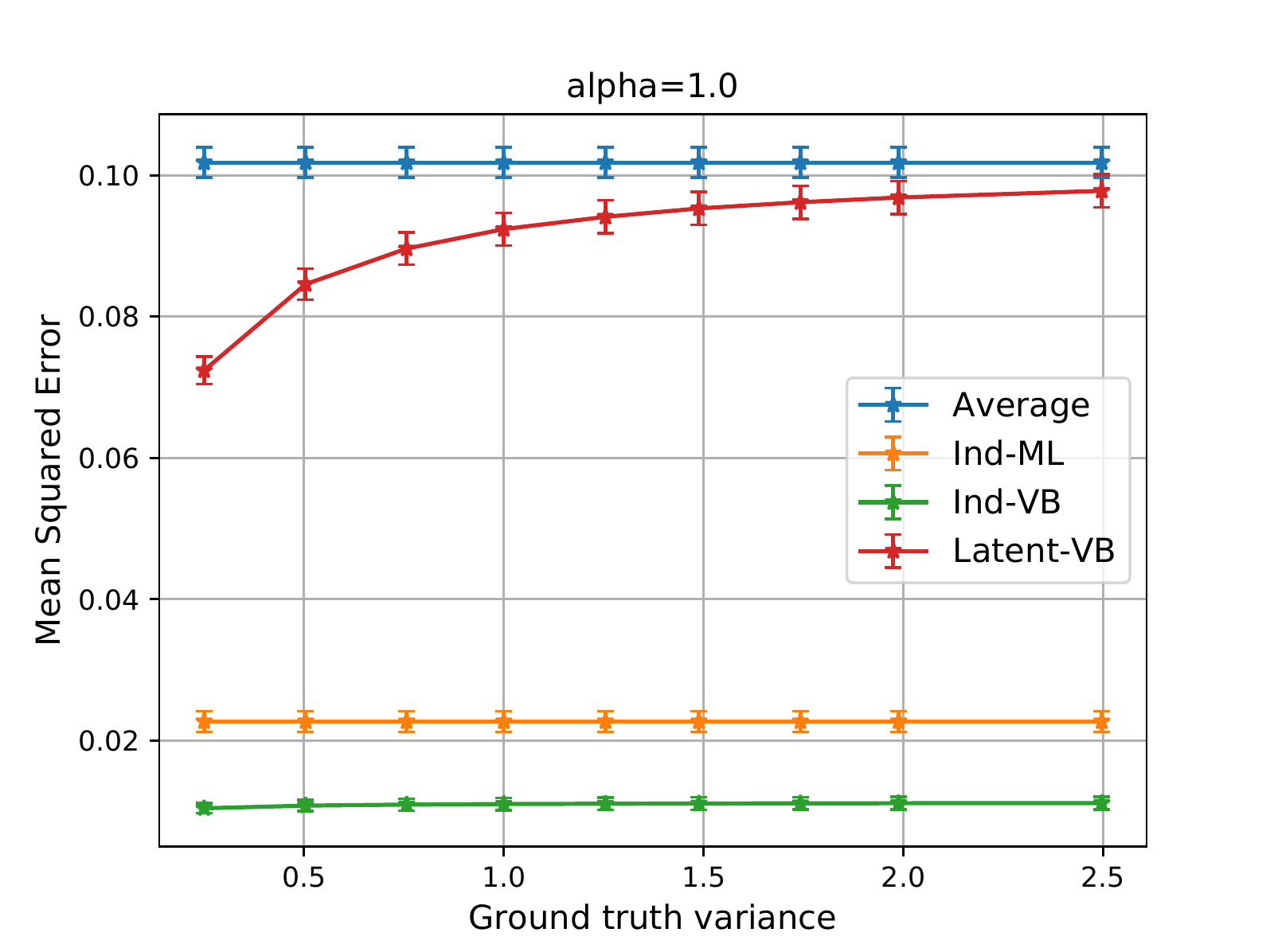} \caption{}
	\end{subfigure} 
	\vspace{-0.1in}
	\caption{Performance on synthetic dataset across different generative models.}
	\vspace{-0.15in}
	\label{fig:syn}
\end{figure*}

We generate synthetic datasets for $I=1000$ crowdsourced prediction tasks and a set of $J = 500$ workers
with $|J_i| = 10$ workers making predictions for each task.
Since our algorithms are based on two different models,
each with  different assumptions on worker noise, 
we generate synthetic data that includes both  components 
and study the robustness of the algorithms to model misspecification.

Specifically, we use:
\begin{align*}
  x_{ij} 
  = \sigma_y y_i + \sqrt{\alpha}\cdot\epsilon_{ij} + \sqrt{1-\alpha}\cdot\xi_{ij},
\end{align*}
where $\epsilon_{ij}$ and $\xi_{ij}$ are 
independent and latent noise components, respectively.
Both components are drawn such that the average variance of each component 
in the system is $1$, and therefore, for all $\alpha\in[0,1]$, 
the total noise variance in the system is $1$. 
The independent noise component is drawn such that 
$\epsilon_{ij} \sim\mathcal{N}(\mathbf{0},\boldsymbol\sigma_j^2)$,
where $\sigma_j^2$ is chosen randomly from the set $\{0.164,1.64,16.4\}$.
The latent noise component is drawn such that $\xi$ is a matrix with rank $D$.
The ground truth values are drawn such that $y_i \sim\mathcal{N}(0,1)$.
Therefore, when $\sigma_y = 1$, the average variance of 
the ground truth and the noise components in the system are equal.

We vary the scaling factor $\sigma_y$ to study 
how the algorithms are affected by 
the noise variance relative to the ground truth variance.
We vary the scaling factor $\alpha$ to study
how the algorithms are affected by
whether the data has a low-rank covariance or diagonal covariance.
All numbers are reported by taking the average of 10 random instances.
Error bars show the 95\% confidence intervals based on standard errors.

Figure~\ref{fig:syn} shows the results of the algorithms' performance on this synthetic crowdsourced prediction dataset.
First, consider the extreme case ($\alpha=1$) where  worker
noise is completely independent across workers. In this case, both
Ind-ML and Ind-VB perform the best, with Ind-VB slightly
better. In the other extreme case ($\alpha=0$) where the
worker noise are completely based on the latent model,
 Latent-VB performs the best.
 
In many cases, we observe that Ind-ML, which is based on
maximizing the likelihood, suffers from overfitting, as can be the case in prediction tasks in general.
In this case, the objective value continues to improve but the actual
performance begins to deteriorate as we update the
parameters. On the other hand, the two approaches based on Variational Bayes are
 robust towards overfitting.

\subsection{Human Age Prediction}
\label{sec:real-data}

We next compare the performance of our proposed algorithms on a real-world crowdsource dataset. The public dataset which is the closest in spirit to our collaborative prediction setting is the human age prediction set
introduced in~\cite{kara2015modeling}. The underlying dataset contains $1002$
pictures of $82$ faces from the $\mathsf{FG}$-$\mathsf{Net}$ database~\cite{FGNet}
in which each photo is labelled with a biological age as the ground truth.
The age of each subject was discretized into $7$ intervals, and $619$
crowdsource workers were asked to predict the age of a subject based on the facial image,
so that each image has $10$ answers from workers.
We shifted all the labels so that the empirical average matches the ground truth
average.

\begin{figure}[!htb]
	\centering
	\includegraphics[width=0.45\textwidth]{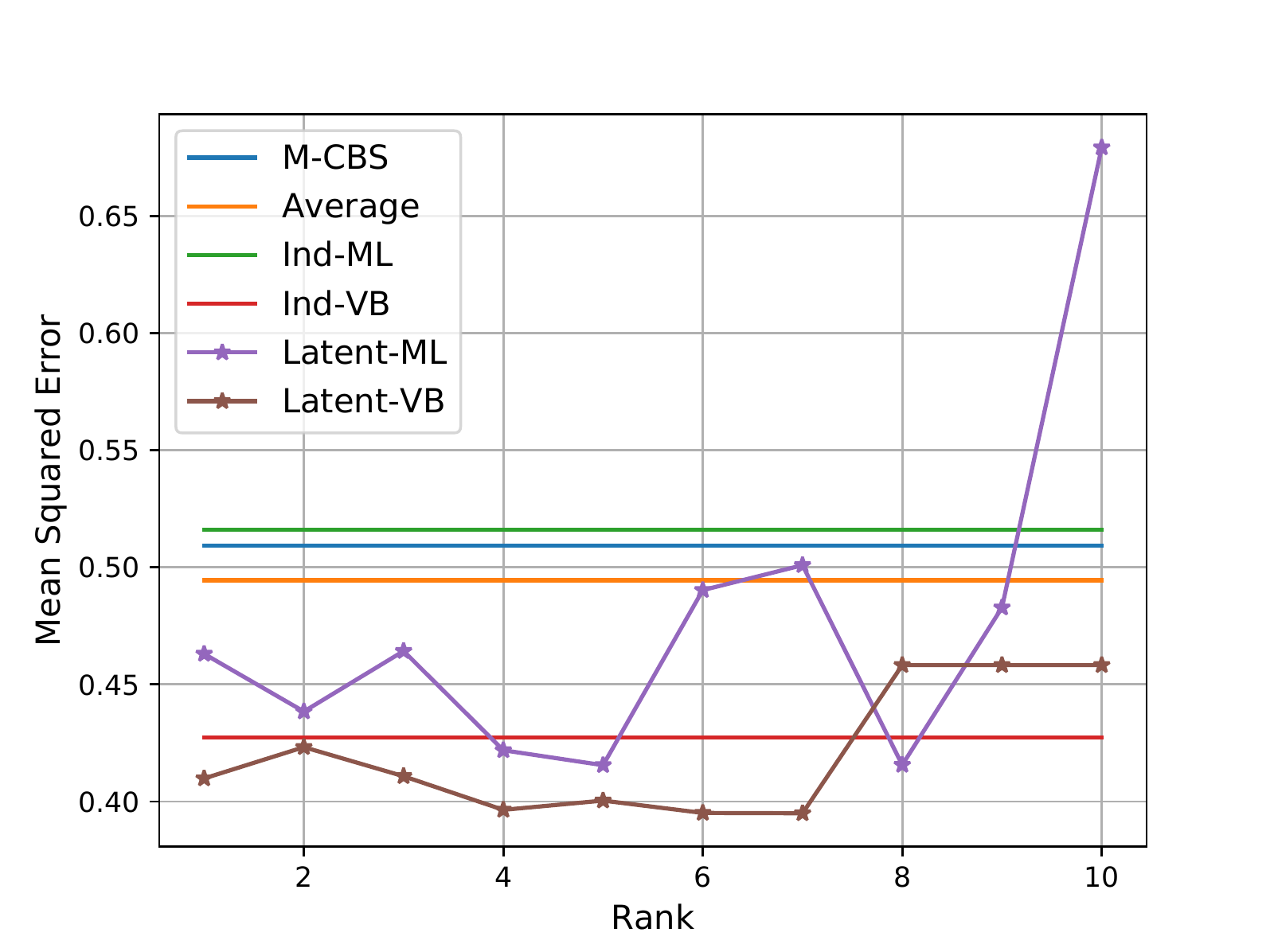}
	\caption{Performance on the age prediction dataset.}
	\label{fig:real}
\end{figure}

Figure~\ref{fig:real} shows the results for the various algorithms. We included the M-CBS
algorithm by~\cite{kara2015modeling} for comparison. For the two approaches based on the latent noise model,
we
tested a range of latent feature dimensions from 1 to 10. Overall, the Variational Bayes approach based on the
latent noise model (Latent-VB) performs the best. We can observe that the method is not
sensitive to the choice of rank. This results suggest that modeling the
correlation between workers can indeed significantly improve the
performance of crowdsourced regression on real data.

\section{Conclusion}
We have introduced a Bayesian framework for the crowdsourcing of predictions or other continuous and regression tasks, when multiple workers perform each task. A standard approach to aggregating the results of the workers over each prediction or regression task makes use of the inverse-variance weight of each worker. We show that using our Bayesian framework we solve two issues that arise in this setting, namely the typical overfitting that occurs in the regression output, and the fact that correlations across workers' responses can influence negatively the results. We demonstrate our proposed approach on a synthetic collaborative prediction dataset as well as on a real publicly available dataset, namely the age prediction crowdsourcing set of \cite{FGNet}.
This approach can help in facilitating the use of crowdsourcing to more complex tasks including in the domain of continuous labels, and function estimation such as using prediction and regression models.

\bibliographystyle{IEEETran}
\bibliography{cdc20}

\end{document}